 \documentclass[final,3p,times,twocolumn]{elsarticle}

\usepackage{amssymb}

\usepackage{longtable}

\journal{ }

\begin{document}

\begin{frontmatter}



\title{A Survey on Predictive Maintenance for Industry 4.0}


\author[WU]{Christian Krupitzer}
\ead{christian.krupitzer@uni-wuerzburg.de}
\author[MA]{Tim Wagenhals}
\author[WU]{Marwin Z\"ufle}
\author[WU]{Veronika Lesch}
\author[Syn]{Dominik Sch\"afer}
\author[Moz]{Amin Mozaffarin}
\author[MA]{Janick Edinger}
\author[MA]{Christian Becker}
\author[WU]{Samuel Kounev}


\address[WU]{Software Engineering Group, University of W\"urzburg, W\"urzburg, Germany}
\address[MA]{Chair of Information Systems II, University of Mannheim, Mannheim, Germany}
\address[Syn]{Syntax Systems GmbH, Weinheim, Germany}
\address[Moz]{MOZYS Engineering GmbH, W\"urzburg, Germany}

\begin{abstract}
Production issues at Volkswagen in 2016 lead to dramatical losses in sales of up to 400 million Euros per week.
This example shows the huge financial impact of a working production facility for companies.
Especially in the data-driven domains of Industry 4.0 and Industrial IoT with intelligent, connected machines, a conventional, static maintenance schedule seems to be old-fashioned.
In this paper, we present a survey on the current state of the art in predictive maintenance for Industry 4.0.
Based on a structured literate survey, we present a classification of predictive maintenance in the context of Industry 4.0 and discuss recent developments in this area.
\end{abstract}


\end{frontmatter}



\section{Introduction}
\label{chap:introduction}
Maintenance has always been a severe cost driver in the production industry. Studies show that depending on the industry between 15 and 70 percent of total production costs originate from maintenance activities \cite{You2010}. Nevertheless, the majority of the production industry still relies on outdated maintenance policies and focuses on an inefficient run to failure approach or statistical trend driven maintenance intervals \cite{Mobley02}.Thus, leading to reduced production time and product quality as a result of inappropriate maintenance policies. According to \cite{Mobley02} surveys on maintenance show that about 33 cents of every dollar spent on maintenance in the US is wasted because of unnecessary maintenance activities. On the other hand, comprehensive research with regard to modern maintenance policies using modern technologies is conducted in different fields of academics such as computer science, production and artificial intelligence. The usage of well-developed sensors and prognostic techniques allows a relatively reliable prediction of the remaining useful life of plant equipment. This so-called predictive maintenance policy is especially of interest in the environment of Industry 4.0 and severely enhances the efficiency of modern production facilities.  
  
Predictive maintenance is based on the idea that certain characteristics of machinery can be monitored and the gathered date be used to derive an estimation about the remaining useful life of the equipment. Hence, this kind of maintenance policy implicates several important improvements in the manufacturing and maintenance process which can severely reduce production costs \cite{Grall2002}. First, predictive maintenance can reduce the number of unnecessary maintenance activities as it is not based on periodic maintenance intervals bound to average lifetime. Thus, potentially reducing the overall number of maintenance activities over a machine's life. Second, not only can too early maintenance activities be avoided, but also too late activities as equipment might fail before the next periodic maintenance interval since the intervals rely on average lifetime which likely includes significant positive but also negative deviations from the mean. For example, due to the specific structure in which a certain component is deployed in larger machinery. Both, the reduction of unnecessary maintenance as well as the reduction of fatal breakdowns result in increased productivity and reduced production down time. Therefore, depending on the accuracy of the prognostic method applied, predictive maintenance can be considered as an overall improvement of efficiency in contrast to conventional maintenance policies \cite{Nguyen2015},\cite{Yam2001}.

	\subsection{Objective and Approach}
	\label{sec:objectives}
The existing research on predictive maintenance is comprehensive and dates back decades with still hundreds of new published papers annually \cite{Jardine2006}. Nevertheless, as the environment is changing and new technologies become available at a more affordable price, there is still a wide range of potential for new research in the field of predictive maintenance. Especially in the context of the Internet of Things (IoT) and Industry 4.0 the possibilities to integrate predictive maintenance and connect it to other systems of the production process are increasing. In order to identify the potential starting points for further research the objective of the present survey is to structure the complex of themes regarding predictive maintenance and to put the comprehensive amount of research into a transparent and comprehensible framework. 
	
The methodical approach of the the present survey is a Structured Literature Review (SLR) based on the keywords \emph{Predictive Maintenance}. Thereafter, all relevant attributes of the selected papers are captured for constructing the framework.

	\subsection{Structure of the paper}
	\label{sec:structure}
The remainder of the survey is structured as follows: Section \ref{chap:foundations} includes the theoretical principals and foundations of predictive maintenance. Next, Section \ref{chap:procedure} describes the methodical procedure of the literature review and the framework construction. Section \ref{chap:framework} explains in detail the categories and the most important attributes of the predictive maintenance framework. Section \ref{chap:discussion} follows a discussion of the framework. Finally, Section \ref{chap:conclusion} concludes the survey by summarizing the main findings and giving recommendations for future research.

\section{Foundations of Predictive Maintenance}
\label{chap:foundations}

Predictive maintenance is a maintenance policy leading to improved efficiency since it allows an estimation of the remaining useful life of the machinery. This approach is based on condition monitoring ideally conducted by sensors, which allows a continuous monitoring process of relevant machine parameters such as vibration and temperature. However, condition monitoring in isolation cannot be considered predictive as it only allows to identify the parameter changes that occur before a failure, but it does not allow to predict a relatively narrow future period of time in which the parameter changes happen and thus the failure might occur. Thus, it does not enable those responsible to schedule maintenance activities ahead in an efficient way, such as finding the most cost efficient time frame for shutting down the production process for maintenance activities. Consequently, a reliable prognostic technique is necessary to transform the acquired data into valuable information for failure prediction. Hence, prognostic techniques are an important part of predictive maintenance and will be an important component of the predictive maintenance framework in Section~\ref{chap:framework}.

In order to clearly differentiate predictive maintenance from traditional conventional maintenance policies Section \ref{subsec:maintpol} describes the most relevant categories of maintenance policies and their unique characteristics. Finally, Section \ref{subsec:e-maint} briefly explains the modern approach of e-maintenance which integrates predictive maintenance into a broader context of manufacturing processes.

		\subsection{Conventional Maintenance Policies}
		\label{subsec:maintpol}
\begin{figure*}
	\centering
	\includegraphics[width=\textwidth]{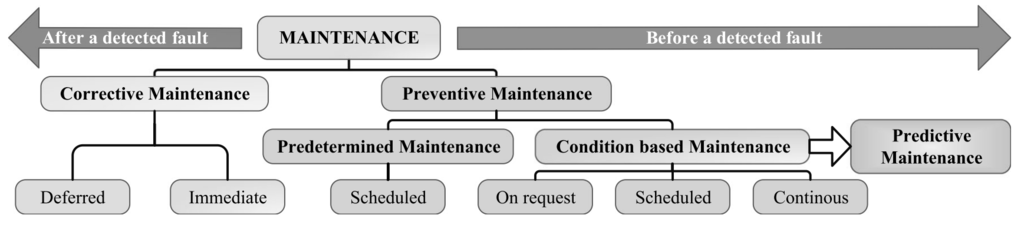}
	\caption{Maintenance Policies \cite{Schmidt2018}}
	\label{img:maintenancepolicies}
\end{figure*}

Generally, maintenance policies can be divided in 2 categories: Corrective maintenance and preventive maintenance. A categorization of conventional maintenance policies is shown in figure \ref{img:maintenancepolicies}. Corrective maintenance always takes place after a failure occurred. Afterwards, the repairing of the machine can be done immediately or at some later point. A production plant that uses this approach is following a \emph{run-to-failure} management based on the philosophy: \emph{If it ain't broke, don't fix it} \cite{Mobley02}. However, this approach can result in severe reduction of manufacturing time ad costly repairs. Corrective maintenance is also referred to as reactive maintenance \cite{Mobley02}. On the other hand, preventive maintenance policies, also called proactive maintenance, are attempted to be executed before a fatal failure occurs. Thereby, preventive maintenance policies are divided into two categories. First, preventive maintenance where the maintenance activities are conducted on pre-scheduled intervals based on historic average equipment lifetime \cite{Schmidt2018}. On the contrary, condition-based maintenance monitors the current condition of a machine and schedules maintenance activities based on the observations made \cite{Schmidt2018}. Here, three distinct condition monitoring methods are feasible: Monitoring on request, scheduled monitoring and continuous monitoring \cite{Grall2002}, \cite{Zhou2007}. The first two methods are mostly based on inspection, while continuous monitoring is generally implemented by sensors. The drawbacks of predetermined maintenance were already discussed in Section \ref{sec:improvements}. The issue with condition-based maintenance is that even with continuous monitoring the acquired data only represents a snapshot of a machines condition. The approach does not allow to efficiently schedule maintenance activities ahead due to missing knowledge about a machine's or component's presumable future state. However, as has already been mentioned and is shown in figure \ref{img:maintenancepolicies} the approach of predictive maintenance is based on the idea of condition monitoring. Hence, predictive maintenance represents an enhancement of mere condition-based maintenance.

		\subsection{E-Maintenance}
		\label{subsec:e-maint}
Mere predictive maintenance is already an improvement in contrast to conventional maintenance policies. Nevertheless, a predictive maintenance strategy which is carried out in isolation of other relevant processes still has potential for optimization. Hence, according to \cite{Lee2006} the efficiency can be severely improved by connecting multiple machines across manufacturing locations of a production facility and monitoring them remotely with wireless sensors, as described in Section \ref{sec:iot}. A web application and cloud-based monitoring infrastructure allows to synchronize the process of maintenance with the overall operation even across production facilities. Furthermore, connected processes such as maintenance resources and even automated spare-part ordering can be integrated in this system \cite{Lee2006}. Precondition for such an e-maintenance system is to aquire and process relevant data in real time. \cite{Muller2008} argue that an e-maintenance can bee seen as an entirely integrated system since it handles the monitoring, diagnosis, prognosis and control process. Thus, e-maintenance goes beyond mere predictive maintenance \cite{Lee2006}. The integrated approach of e-maintenance matches the concept and idea of Industry 4.0~\cite{Yan2017}.

\section{Methodical Procedure}
\label{chap:procedure}
The following section describes the methodology used to select the relevant papers and construct the framework.
In total 150 papers are selected to derive the framework. 
All the attributes are clustered into 10 categories that build the main layer of the framework. 

	\subsection{Structured Literature Review}
	\label{sec:review}
The Structured Literature Review (SLR) is a systematic approach to find the relevant literature in order to answer a set of research questions by searching for papers based on predefined key words \cite{Slr14}. By searching with specific key words the method allows to find the most relevant literature to address the research questions. However, there is no guarantee that a SLR is able to find all relevant literature \cite{Slr14}. Nevertheless, the SLR has the advantage to tackle a specified topic from numerous directions, allowing the researcher to properly cover the important sub themes \cite{Slr14}. Thereby, the SLR allows to uncover gaps in the existing primary research and helps to reveal areas where additional research might be needed \cite{Slr14}.

As the field of predictive maintenance is extremely broad and has been addressed comprehensively by researchers for decades, the existing pool of papers is enormous. Therefore, the only key words used in the present survey to detect the relevant literature with a Google Scholar search are \emph{Predictive Maintenance}. As the objective of the survey is to construct a framework addressing the distinctive facets of predictive maintenance this procedure results in the most favorable set of important literature since a combined search of predictive maintenance and Industry 4.0 mainly results in papers which address Industry 4.0 but only mention predictive maintenance as part of it without explaining a specific approach in detail. Thus, with these papers the development of a detailed framework with all facets of predictive maintenance would not be possible. As already mentioned, the tremendous amount of research addressing the topic of predictive maintenance results in thousands of potentially relevant papers which cannot be evaluated in a single survey. Thus, the present survey does not claim to be entirely complete and only represents an overview. The search algorithm of Google Scholar runs very successful as the first papers of the search are the most suitable papers for the topic, while the quality and usefulness of the later papers is steadily decreasing. This allows to identify appropriate literature by checking the list of results from the top down. The search for academic papers via Google Scholar was conducted in October 2018. A paper for the framework was selected by first scanning the abstract to identify the detail in which the topic of predictive maintenance is covered in the paper. If a paper is explaining the applied predictive maintenance approach and its context, e.g. system size and condition monitoring in detail, it is utilized to build the framework.
The selected papers are published between 1993 and 2018 while almost two thirds (97/150) are published during the last decade. Since the research concerning predictive maintenance is a very broad field it is addressed by various types of journals and conferences.

Another literature review method that was considered for the present survey is the so-called berry picking method. In this method the review starts by identifying a starting paper that matches the addressed topic and objectives \cite{berry08}. The following step in the process is the footnote chasing where the list of references of the starting paper is checked for more relevant literature \cite{berry08}. Furthermore, the search for relevant literature can be extended by checking not only the references of the starting paper, but also checking the papers which cite the starting paper \cite{berry08}. The drawback of this method is that the papers are most likely connected based on a similar sub theme that is addressed by these papers, thereby potentially missing important other subsections of the main topic. Of course, this issue can be approached by choosing multiple starting papers to cover the missing sub themes. Nevertheless, the amount of literature addressing predictive maintenance is so broad and contains numerous sub themes that even with an increased number of starting papers the berry picking method would still miss relevant fields. Hence, the SLR is considered as better method for the present survey as it allows to find more diversified relevant literature in order to cover a much broader scope of the topic. Thus, resulting in a more favorable result for building the framework and addressing the first research question specified in section \ref{sec:objectives}.

	\subsection{Construction of the Framework}
	\label{sec:frameworkconstruction}
The construction of the framework starts by examining the first paper that was selected from the top down of the potentially relevant literature of the Google Scholar search. Notice that it is not important with which paper this process is started as the whole process has an iterative characteristic. While a paper is examined a table is filled with data about the relevant attributes covered in the paper. Thus, whenever an important attribute or characteristic concerning the topic of predictive maintenance is identified a new column is added to the table and the attribute is ticked for this paper. Thereafter, every paper that is categorized in the table gets a separate line to mark the attributes which are covered in this paper. Thereby, the following papers are checked iteratively as their research content is searched for the already existing attributes. Additionally, the following papers are checked for additional relevant attributes which are added to the table. The data grid that is generated by this process represents the foundation of the predictive maintenance framework. It is important to mention that an attribute is merely considered as covered by a paper if it is a direct research subject of the paper. Hence, a literature review within a primary study is not considered sufficient to fulfill certain attributes. However, the list of categorized papers also contains 10 surveys (\cite{Jardine2006}, \cite{Lee2006}, \cite{DeFaria2015}, \cite{Efthymiou2012}, \cite{Edwards1998}, \cite{Heng2009}, \cite{Kandukuri2016}, \cite{Lu2009}, \cite{Munirathinam2014}, \cite{Muller2008}) where the approach is obviously different and all referenced attributes are marked in the table.

In the next step the raw data grid is structured by grouping attributes which are similar to each other or belong to the same sub theme. Subsequently, each cluster is named by a term which represents the content of the cluster. These umbrella terms build the main categories of the framework. If appropriate, the clusters can have further subdivisions to allow for a finer classification within the categories. Thus, the entire framework can be best represented in form of a tree. The main categories of the predictive maintenance framework, as well as the tree structure for every category are presented in Section~\ref{chap:framework}.

\section{Framework for Predictive Maintenance}
\label{chap:framework}

\begin{figure}[h]
	\centering
	\includegraphics[width=\columnwidth]{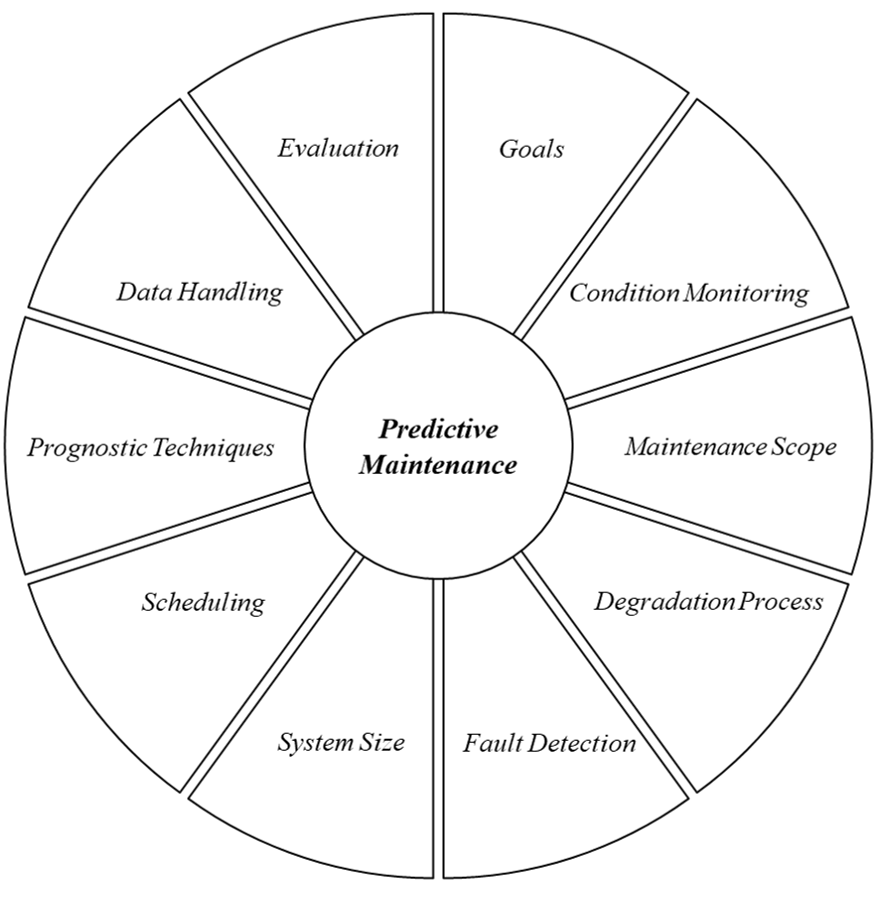}
	\caption{Categories of the Framework for Predictive Maintenance}
	\label{img:framework}
\end{figure}

In section \ref{sec:frameworkconstruction} the methodical approach for structuring the framework was explained. The analysis of the papers resulted in a grid which consists of a total of 69 different attributes. The framework that is built from this data grid is described in the following sections. The entire framework for predictive maintenance consists of 10 categories. These categories are Goals, Condition Monitoring, Maintenance Scope, Degradation Process, Fault Detection, System Size, Scheduling, Prognostic Techniques, Data Handling and Evaluation. Figure \ref{img:framework} shows these 10 categories which represent the highest level of the framework. The following sections of describe the categories in detail and show all the attributes which are summarized within a certain category. 

	\subsection{Goals}
	\label{sec:goals}

\begin{figure}[h]
	\centering
	\includegraphics[width=\columnwidth]{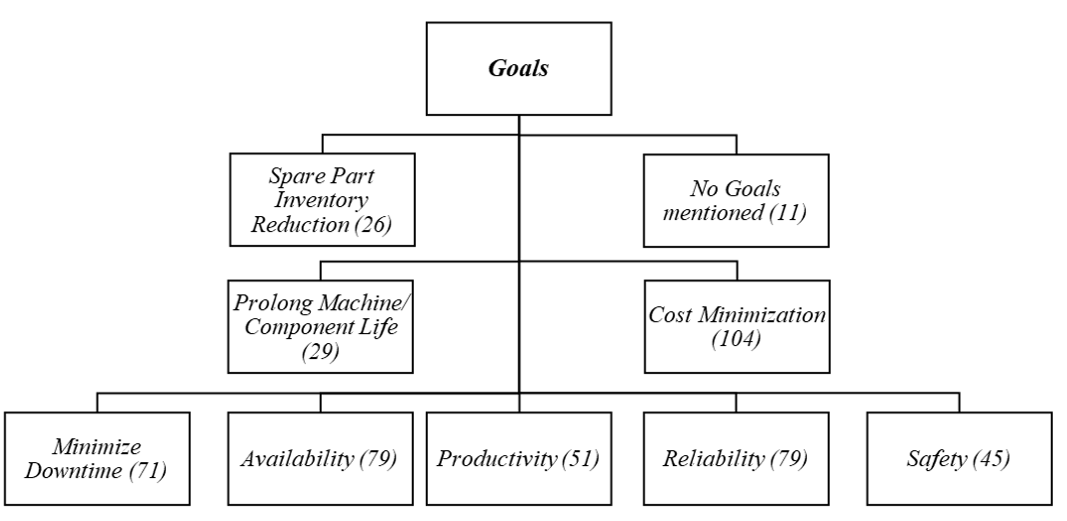}
	\caption{Overview of the Category Goals}
	\label{img:goals}
\end{figure}

The category of goals is generally used by researches to support their motivation for choosing the research topic of predictive maintenance. Thus, 139 out of 150 papers (93\%) mention one or several goals of the framework. During the analysis of the paper 8 goals have been identified: Spare Part Inventory Reduction, Prolong Machine/Component Life, Cost Minimization, Minimize Downtime, Availability, Productivity, Reliability and Safety. Some of these goals were already introduced in section \ref{sec:improvements} but not explained in detail. Figure \ref{img:goals} shows all attributes belonging to the category of goals. The number in parentheses is the frequency of occurrence of a certain attribute out of a total of 150 papers. 

First, predictive maintenance has the potential to reduce the number of spare parts in stock since a reliable predictive maintenance system would allow for the possibility to merely store the spare parts needed soon instead of storing every spare part which could potentially be necessary \cite{Zhou2007}. Hence, the predictive maintenance system would reduce the number of spare parts in stock and the overall storage size while still maintaining a proper maintenance process and avoiding production downtime due to unavailable spare parts.   

Second, some researchers argue that predictive maintenance can in fact prolong the overall life of machines and components \cite{Wu2007}. This is because the predictive maintenance system monitors the health condition and predicts the remaining useful life of a machine and thus reduces the risk of a fatal breakdown which might reduce the lifespan of the machine even in case of a thoroughly corrective maintenance action \cite{Wu2007}. Furthermore, as predictive maintenance systems also help to avoid unnecessary maintenance actions this can additionally prolong a machine's life because every maintenance action puts the machine to risk since a maintenance action and replaced components represent an intervention in the structure of a machine \cite{Baidya2015}.

The by far most frequently mentioned goal is the objective to minimize costs through a predictive maintenance approach \cite{Deloux2009}. More than two thirds of the analyzed papers mention cost minimization as one of the main goals of predictive maintenance and thus their research. The goal of cost minimization is often related to other goals of a predictive maintenance system such as minimizing downtime by avoiding fatal breakdowns or prolong machine life and reduce the need for spare part inventory. As a reliable predictive maintenance system leads to an overall more efficient way of the maintenance process this results in an overall reduction of costs. Hence, while the implementation of a predictive maintenance system is costly and complex it is argued to be a positive business case in the long run \cite{Deloux2009}.

Fourth, since predictive maintenance systems allow fault detection in the future based on monitoring data and prognostic techniques the number of fatal breakdowns can be reduced \cite{Kaiser2009}. With a conventional scheduled maintenance policy based on historic data about the life span of a machine or component, there is still the possibility that a machine or component might fail before the next maintenance interval since the interval is merely an average that includes outliers. In contrast to that, a predictive maintenance approach monitors the machine and its components regularly or even continuously and triggers maintenance actions regardless of the time a machine or component is in use \cite{Mobley02}. Additionally, a predictive maintenance system can also minimize the downtime of a manufacturing line as it allows to plan maintenance actions ahead and group certain maintenance actions to reduce the number of production stops for single maintenance actions \cite{Yang2008}. Minimizing downtime has a severe effect on reducing costs and increasing productivity. Thus, almost half (47\%) of the papers in this survey address the objective of minimizing downtime.  

The goals of availability, productivity and reliability are all connected to the goals of cost minimization and minimizing downtime \cite{Raza2017}, \cite{Okoh2017}, \cite{Baidya2015}. By decreasing downtime a predictive maintenance system automatically increases the total amount of time a machine can be in an online state. The increased overall manufacturing time also increases the productivity of a manufacturing plant as more products can be produced in the same time while the monitoring component of the predictive maintenance system always makes sure that the equipment is running in a healthy condition. Hence, the predictive maintenance approach keeps machinery in a state where it produces products in a good quality \cite{Carnero2006}. Thus, this leads to an increase in productivity. The goal to increase the reliability of machines is strongly related to the goal availability as these two objectives are always addressed together. Additionally, availability, productivity and reliability increase the output and quality in a certain time interval and therefore minimize production costs.

Finally, safety is a goal that is not directly related to business performance indicators such as productivity and costs. Nevertheless, it is still an important objective in the context of predictive maintenance research since it is addressed by almost a third (30\%) of all papers in the present survey. At present, most production lines still require the interaction of humans and are not fully automated. Therefore, humans work nearby or directly with heavy machinery. Fatal breakdowns and unreliable running conditions of these machines cause a potential risk to the employees. Hence, as predictive maintenance avoids fatal breakdowns and monitors the health condition of machines, the safety of employees is improved \cite{Efthymiou2012}.

	\subsection{Condition Monitoring}
	\label{sec:conditionmonitoring}	

\begin{figure}[h]
	\centering
	\includegraphics[width=\columnwidth]{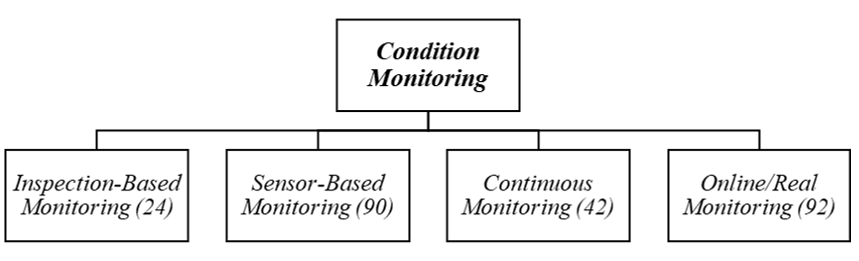}
	\caption{Overview of the Category Condition Monitoring}
	\label{img:conditionmonitoring}
\end{figure}

A predictive maintenance approach is based on the collection of data from a machine or component which indicates its health status and allows the prediction of the residual useful life based on this monitoring data \cite{Gebraeel2005}. The analysis of the 150 papers has identified 4 different attributes for the category of condition monitoring: Inspection-Based Monitoring, Sensor-Based Monitoring, Continuous Monitoring and Online/Real Monitoring. All attributes and their corresponding frequency of occurrence are shown in figure \ref{img:conditionmonitoring}. 

First, the least addressed approach to monitor a machine's condition and collect valuable data is inspection-based monitoring (16\%). With inspection-based monitoring the data is merely collected in inspection intervals. However, the intervals are not predefined as it is the case in conventional maintenance policies. The intervals are adapted with respect to the observed and collected data about a machine's or component's current and predicted conditional state \cite{Jardine2006}. 

Second, since advancements in sensor technology make sensors for various types of parameters more affordable, the majority of studies base their research on sensor-based monitoring. With sensor-based monitoring different types of sensors for instance to observe vibration and temperature are used to collect the relevant data \cite{Orhan2006}. Note that sensor-based monitoring and inspection-based monitoring are not mutually exclusive as it is often the case that sensor equipment is necessary to perform the inspection \cite{Kaiser2009}. Generally, the usage of sensor-technology is more appropriate for an integrated predictive maintenance system as it is crucial for efficient continuous monitoring.

As the term already indicates, continuous monitoring is the continual collection of relevant monitoring data to estimate the remaining useful life of a machine or component \cite{Traore2015}. In contrast to inspection-based monitoring the amount of data collected is significantly higher since inspection-based monitoring is merely a periodic snapshot of a machine's conditional state. Hence, inspection-based monitoring and continuous monitoring are the only two attributes in this category which are mutually exclusive. All other attribute combinations are feasible.    

Finally, online/real monitoring is a condition monitoring technique which allows the collection of data in the running state of a machine and is addressed by the majority of the analyzed papers (61\%) \cite{Lindstrom2017}. Furthermore, it is the prerequisite for continuous monitoring as a continual collection of data is merely feasible in the running state of a machine. Therefore, researchers always address online monitoring when implementing a continuous monitoring approach in their studies. However, online monitoring is also possible for inspection-based methods, but in this case it does not resemble a requirement.

	\subsection{Maintenance Scope}
	\label{sec:maintenancescope}

\begin{figure}[h]
	\centering
	\includegraphics[width=\columnwidth]{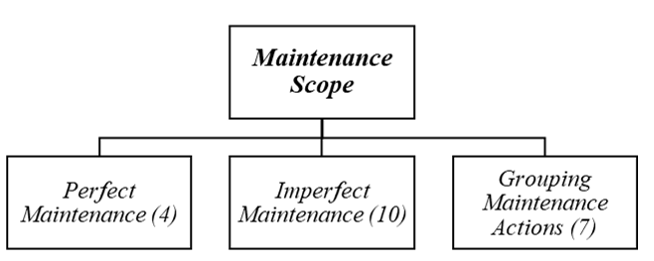}
	\caption{Overview of the Category Maintenance Scope}
	\label{img:maintenancescope}
\end{figure}

The category of maintenance scope is a category which is generally not addressed frequently. Obviously, for every maintenance action there has to be an assumption about the maintenance scope but merely a few studies (13\%) mention this category directly. Therefore, the framework solely represents the number of times the topic of maintenance scope is addressed directly. The present survey makes no assumptions about the maintenance scope in cases where no maintenance scope is mentioned directly. The analysis of the papers revealed 3 different attributes for the category maintenance scope: Perfect Maintenance, Imperfect Maintenance and Grouping Maintenance Actions. Figure \ref{img:maintenancescope} shows all attributes belonging to the category of maintenance scope. 

First, the attribute of perfect maintenance is based on the assumption that every maintenance action conducted at a machine or component restores the functionality and durability to its original level. Hence, into the condition of a new machine \cite{Dieulle2001}. This assumption follows an \emph{as good as new} approach for every maintenance action \cite{Dieulle2001}. In contrast to this approach, the assumption of imperfect maintenance is based on the premise that a maintenance action cannot restore the functionality and durability of a certain machine into an \emph{as good as new} state, but only into an \emph{as good as old} condition \cite{Tan2010}. Thus, the machine or component is still assumed to be in the state of used equipment even after maintenance. More studies make the assumption of imperfect maintenance compared to perfect maintenance since they argue that it is a more realistic conjecture \cite{Zhou2007}. Generally, the category of maintenance scope is not very widespread in the predictive maintenance research, presumably because the usually continuous monitoring and prediction updates make the assumption about the maintenance scope obsolete.

The third and final attribute of the category of maintenance scope seems to be far more important for predictive maintenance but is as well not severely represented in the academic literature. The possibility to group maintenance actions in an efficient way leads to an overall cost reduction for maintenance activities as downtime can be reduced \cite{Ladj2016}, \cite{Nguyen2017}. Precondition for grouping maintenance actions is a holistic predictive maintenance approach which monitors the entire manufacturing equipment simultaneously in order to identify certain maintenance actions which are best conducted at the same time \cite{Nguyen2017}. While the assumptions about perfect and imperfect maintenance are mutually exclusive, the attribute of grouping maintenance actions can be addressed in combination with the other two attributes of the category. The issue that merely a few of the analyzed papers address the important attribute of grouping maintenance will be discussed in Section \ref{chap:discussion}.

	\subsection{Degradation Process}
	\label{sec:degradationprocess}

\begin{figure}[h]
	\centering
	\includegraphics[width=\columnwidth]{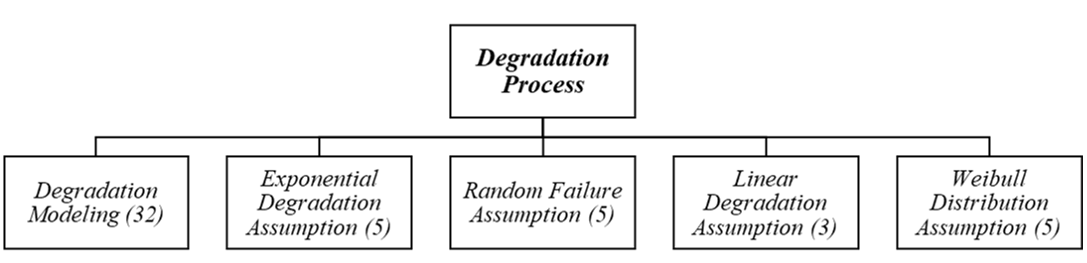}
	\caption{Overview of the Category Degradation Process}
	\label{img:degradationprocess}
\end{figure}

The category degradation process is addressed by a third (34\%) of the analyzed papers and covers either the direct modeling of the degradation process of a machine or a predefined assumption about its deterioration course. The attributes identified for this category are: Degradation Modeling, Exponential Degradation Assumption, Random Failure Assumption, Linear Degradation Assumption and Weibull Distribution Assumption. Figure \ref{img:degradationprocess} shows the category degradation process and all its attributes. 

\begin{figure}
	\centering
	\includegraphics[width=\columnwidth]{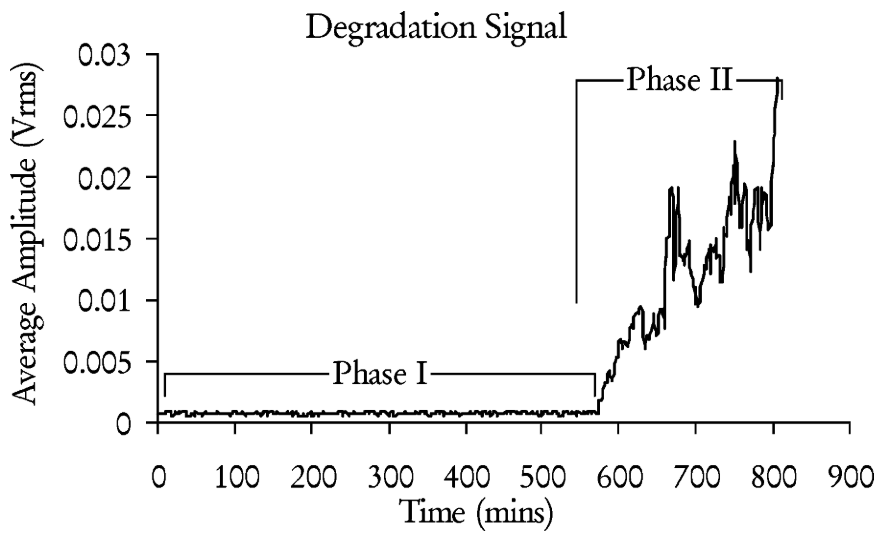}
	\caption{Vibration-based Degradation Modeling \cite{Gebraeel2006}}
	\label{img:degradationmodeling}
\end{figure}

First, the process of degradation modeling is the derivation of the deterioration course of a machine or component based on relevant machine health indication data such as vibration or temperature \cite{Gebraeel2006}. The reason for this procedure is to gather information about a certain machines typical conditional state over its lifetime \cite{Gebraeel2006}. The modeling of the deterioration process is beneficial in a sense that having knowledge about a machine's degradation pattern can support a predictive maintenance system to predict future breakdowns more accurately. Nevertheless, the predictive maintenance approach is not meant to merely rely on the average degradation of a machine to decide on the maintenance intervals. The main indicator is still the predictive maintenance system and its prognostic approach that could be supported by the information of the degradation modeling. Figure \ref{img:degradationmodeling} shows an example for an vibration-based degradation course where phase I represents the non-defective state and phase II the conditional state close to failure \cite{Gebraeel2006}. Thus, the objective is to model these phases in order to support the machinery prognostics of the predictive maintenance decision with regard to the timing and need for maintenance activities.

A few papers make an assumption about the degradation process instead of following a modeling approach. However, since the predictive maintenance approach is based on monitoring the conditional state and prognostics, the mere assumption about the deterioration course of a machine is not really relevant for a predictive maintenance approach. Hence, only a fraction of the academic literature mentions such an assumption prior to the implementation of an predictive maintenance approach. The assumptions about linear degradation, exponential degradation and random failure are straight forward \cite{Elwany2008}, \cite{Gebraeel2005}, \cite{Hashemian2011a}. A predictive maintenance approach would be most efficient in the prsence of random failure since conventional maintenance policies would fail most of the time. Additionally, the academic literature mentions a distribution called Weibull Distribution \cite{Gebraeel2005}. It is a continuous probability distribution based on adjustable parameters which is used to model the lifespan of machines or components \cite{Gebraeel2005}.

	\subsection{Fault Detection}
	\label{sec:faultdetection}	
	
\begin{figure}[h]
	\centering
	\includegraphics[width=\columnwidth]{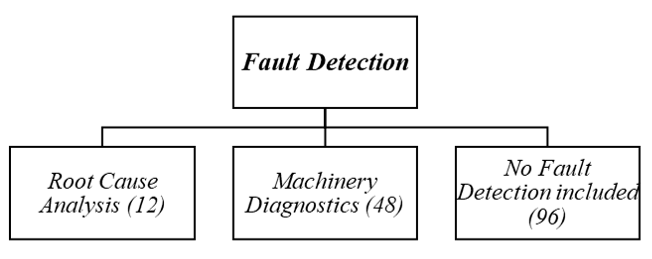}
	\caption{Overview of the Category Fault Detection}
	\label{img:faultdetection}
\end{figure}

A pure predictive maintenance approach solely focuses on the prediction of the future conditional state of machinery and components to schedule maintenance activities in an appropriate way and scope.  Nevertheless, a little bit over a third (36\%) of the examined academic papers additionally address the topic of fault detection, meaning that the predictive maintenance approach does not only attempt to predict the remaining useful life of the machine, but also tries to identify the root cause of the failure based on the collected data \cite{Yam2001}. The category of fault detection includes the following attributes: Root Cause Analysis, Machinery Diagnostics and No Fault Detection. All attributes are shown in figure \ref{img:faultdetection}. 

Fault detection covers the additional function of diagnostics. Thereby, root cause analysis and machinery diagnostics address the same issue. While most researchers call it machinery diagnostics, some also refer to it as root cause analysis or mention both terms. The general idea is the processing of acquired monitoring data to uncover the reasons for future failure. Thus, using vibration or other machine monitoring data for diagnostic purposes \cite{DeFaria2015}. The feasibility and accuracy of a fault detection approach depends on the level of monitoring activity which means that the more machine parts and components are monitored separately, the better can be identified where the root cause for a future failure may lie \cite{DeFaria2015}.

	\subsection{System Size}
	\label{sec:systemsize}		
	
\begin{figure}[h]
	\centering
	\includegraphics[width=\columnwidth]{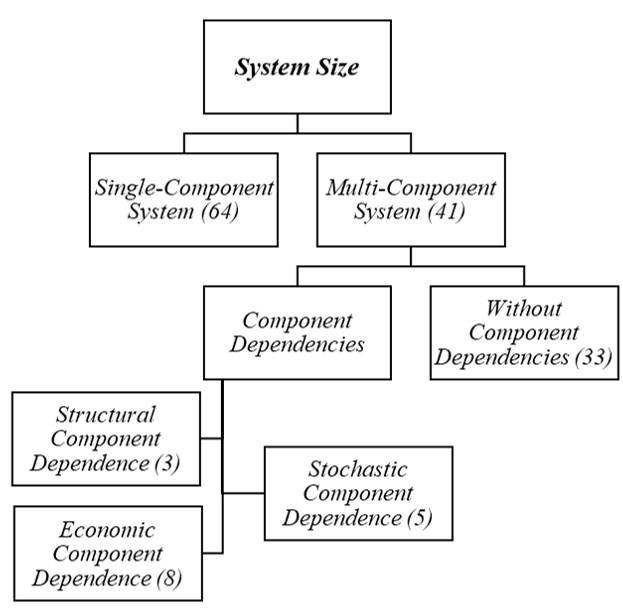}
	\caption{Overview of the Category System Size}
	\label{img:systemsize}
\end{figure}

Another relevant parameter which was identified during the literature review is the category of system size. This category addresses the extent to which the predictive maintenance approach is applied or assumed to be applied when implemented in real life. The analysis revealed two attributes: Single-Component Systems and Multi-Component Systems. Furthermore, the attribute of multi-component systems is further divided in the sub category of component dependencies which are divided into the following attributes: Structural Component Dependence, Economic Component Dependence and Stochastic Component Dependence. Figure \ref{img:systemsize} shows all attributes of the category. A little bit over two thirds (70\%) of the papers analyzed in the present survey address the category of system size. Hence, this category can be seen as important in the context of predictive maintenance.

First, single-component systems are defined by the present survey as either single components, e.g. experimental studies which conduct laboratory tests with merely single bearings, or single machines considered solely in an isolated context \cite{Hashemian2011}. This means that the machine itself might consist of multiple components but the total machine as a whole is not considered as a true multi-component system. The definition chosen for this category results in less true multi-component systems compared to single-component systems (61\% vs. 39\%).

As already described in the previous paragraph, the true multi-component system consists of multiple machines or separate components which together form or are part of larger system, e.g. an entire manufacturing line \cite{VanHorenbeek2013}. The implementation of an successful predictive maintenance system is much more complicated for these multi-component systems since more data needs to be processed and dependencies between the system's components become relevant \cite{VanHorenbeek2013}.  However, while there is a significant number of papers addressing multi-component systems, merely 8 of the 41 papers (20\%) additionally address the topic of dependence. 

Dependence in multi-component systems can be divided in three different types. First, economic dependencies are such dependencies that enable cost reduction when parts of the system are maintained simultaneously, e.g. because for the maintenance of one component other components have to be offline as well, thereby reducing downtime when maintenance actions for these components are conducted jointly \cite{Nguyen2015}. Second, stochastic dependencies are dependencies in consequence of stochastic relations between components of their deterioration process. Hence, the degradation of one component affects the state of one or multiple other components of the system \cite{Nguyen2015}. Finally, structural dependencies result from components which form a unified part in a sense that the maintenance of one component directly implies the maintenance of all structural dependent components \cite{Nguyen2015}. The most important of these three types of dependencies seems to be the economic dependence as all papers which cover the topic of dependence at least mention this type. Furthermore, the existence of these dependencies is the reason why single-component predictive maintenance approaches are not simply scalable to a multi-component level but must be adapted with regard to the affects of these dependencies \cite{Nguyen2015}.

	\subsection{Scheduling}
	\label{sec:scheduling}

\begin{figure}[h]
	\centering
	\includegraphics[width=\columnwidth]{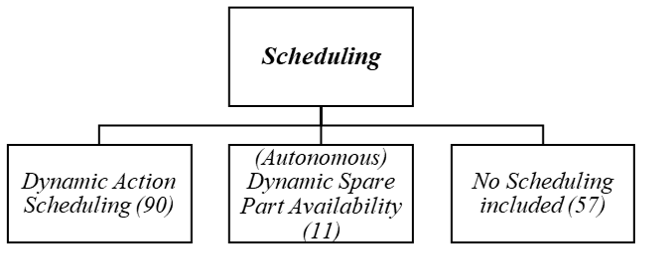}
	\caption{Overview of the Category Scheduling}
	\label{img:scheduling}
\end{figure}

The category of scheduling is addressed by almost two third (62\%) of the papers and therefore has an important role in the field of predictive maintenance. This does not seem to be unusual as the main motivation behind a predictive maintenance approach is to identify the need and timing for maintenance activities in advance, allowing for an efficient scheduling of these activities. The attributes identified in this category are: Dynamic Action Scheduling, (Autonomous) Dynamic Spare Part Availability and No Scheduling Included. Figure \ref{img:scheduling} shows all attributes of the category and their corresponding frequency of occurrence.

First, dynamic action scheduling describes the possibility to dynamically adapt the maintenance schedule based on new and processed condition monitoring data \cite{Yang2008}. This dynamic scheduling is only possible in a predictive maintenance environment due to the forecast of a machine's future conditional state. Optimization algorithms can be applied to define the most cost-effective maintenance schedule and continuously update this schedule when new machinery prognostics information becomes available \cite{Yang2008}. Hence, the maintenance schedule is not static as it would be for conventional maintenance policies but rather dynamic. 

Additionally, a few papers cover the attribute of (autonomous) dynamic spare part availability, where not only the maintenance activities itself but also the necessary spare part ordering is linked to the predictive maintenance system \cite{Nguyen2017}. This broader and more integrated approach better fits a modern idea of Industry 4.0 and smart factory compared to an isolated policy. However, over a third (38\%) of the analyzed academic literature does not address the category of scheduling in conjunction with predictive maintenance  directly.

	\subsection{Prognostic Techniques}
	\label{sec:prognostictechniques}

\begin{figure}[h]
	\centering
	\includegraphics[width=\columnwidth]{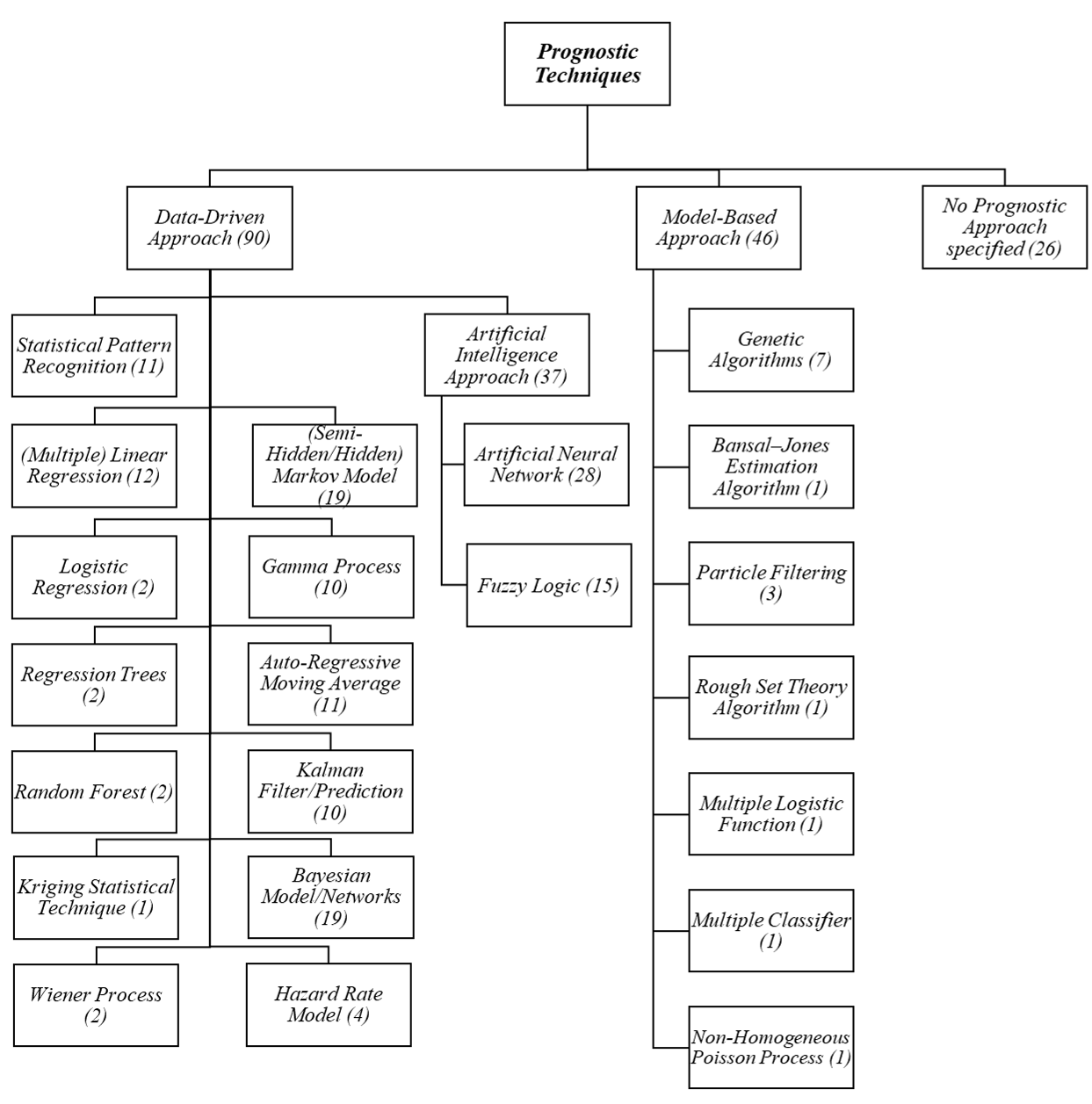}
	\caption{Overview of the Category Prognostic Techniques}
	\label{img:prognostictechniques}
\end{figure}

The category of prognostic techniques is clearly one of the most important ones for predictive maintenance. While all the monitoring and data acquisition is indispensable the prognostic technique is what transforms the raw data into valuable information. Note that since a prognostic technique attempts to predict a prospective failure of a machine or component the generated information are just probabilities. For the predictive maintenance framework of the present survey 3 attributes were identified for the category of prognostic techniques: Data-Driven Approach, Model-Based Approach and No Prognostic Approach specified. Figure \ref{img:prognostictechniques} shows all 3 primary attributes as well as their secondary attributes and their corresponding frequency of occurrence.

The analysis of the 150 papers showed that the number and diversity of different techniques is enormous. The majority of techniques is often found only once or twice within all the papers. Thus, the present survey will not present and discuss the different techniques in detail. In addition, the attribute no prognostic approach specified is added to the predictive maintenance framework since not every paper mentions the applied approach. For the cases in which the papers mention their approach there is no single definition of how to classify these techniques into the attributes data-driven and model-based. Furthermore, some papers just mention that they apply some kind of model-based or data-driven approach but do not specify this approach further, which makes it difficult to stick to one definition for all papers.  

It is important to take into consideration that all these prognostic techniques rely on monitoring data for their predictions. Hence, in a broad sense all these methods are somehow data-driven. However, in the present survey data-driven approaches are such techniques which rely heavily on previous data input in order to enable reliable predictions in the first place \cite{Yuan2013}, \cite{Datadriven11}. Good examples are (multiple) linear regression, logistic regression, regression trees, random forest and the kriging statistical technique. All these techniques are based on the concept of regression \cite{Datadriven11}. However, for a reliable regression model there must be a huge amount of training data to define the model before it can actually be applied to new test data. The same holds for bayesian-based models. Prior to work with the bayesian principle of believe updates the a priori probabilities must be derived from training data \cite{Datadriven11}. 

Beside these classical statistic methods the attribute data-driven approach also has a sub attribute category of prognostic techniques related to artificial intelligence. Almost 25\% of the papers mention an artificial intelligence approach. Most frequently the concept of artificial neural networks. Artificial intelligence concepts also belong to the group of data-driven approaches since they require an enormous amount of training data, too \cite{Kandukuri2016}, \cite{Wu2007}. In the case of an artificial neural network nodes which are also called neurons are structured in multiple layers where every neuron passes on a value to all nodes in the next layer \cite{Ann00}, \cite{Ann02}. Every value is weighted by some real number that represents the weight of the connection between two neurons \cite{Ann00}, \cite{Ann02}. The idea of the network is that a specific input results in a specific outcome with a certain probability \cite{Ann00}, \cite{Ann02}. The network mostly depends on the weights of every connection. However, finding these weights is no easy task and requires a huge amount of training data in order to build a reliable artificial neural network \cite{Ann00}, \cite{Ann02}. In total 73\% of the 124 papers which mention a prognostic technique, mention a data-driven approach.

As already mentioned, the difference between data-driven and model-based approaches is not straightforward but rather vague. Generally, a data-driven prognostic technique is based on a large amount of available historic monitoring data while a model-based technique is applied for new or untested systems which lack this comprehensive measurement data \cite{Yuan2013}. Furthermore, about 8\% of the papers mention both the data-driven and the model-based approach. In the predictive maintenance framework of the present survey model-based approaches are defined as techniques which process monitoring data but do not require a great amount of training data before they can be applied. However, the allocation of techniques into this attribute is vague and often based merely on the fact that a method does not fit the definition of a data-driven approach properly. Furthermore, the framework shows that about a third (31\%) of the papers apply a model-based approach but the frequencies of the sub attributes are very low. This is the case since a lot of the papers in the model-based attribute merely mention the usage of some type of model for the machinery prognostics part. Thus, even more making the allocation within this framework category a vague process. However, the main purpose of this category is not to find a generally valid classification of every single prognostic technique as this is not feasible, but to demonstrate the variety and number of techniques identified after analyzing just 150 papers of the research area of predictive maintenance.

	\subsection{Data Handling}
	\label{sec:datahandling}

\begin{figure}[h]
	\centering
	\includegraphics[width=\columnwidth]{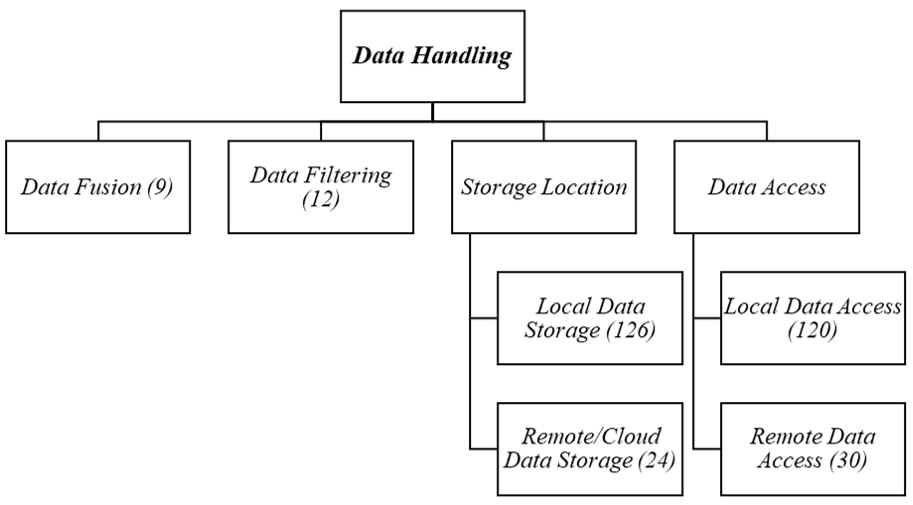}
	\caption{Overview of the Category Data Handling}
	\label{img:datahandling}
\end{figure}

The category data handling deals with the amount of data acquired by condition monitoring. Especially in cases of continuous monitoring with multiple sensors the amount of data collected by these sensors results in an enormous amount of data to be handled \cite{Munirathinam2014}, \cite{Yan2017}. The analysis of the 150 papers covered in the present survey reveals the following relevant attributes: Data Fusion, Data Filtering, Storage Location and Data Access. Additionally, the attributes of storage location and data access are divided into the sub attributes of Local Date Storage and Remote/Cloud Data Storage as well as Local Data Access and Remote Data Access. Figure \ref{img:datahandling} shows all attributes of the category and their corresponding frequency of occurrence. 

First, data fusion and data filtering are two methods which deal with the issue of big data generated by continuous monitoring. The approach of data fusion utilizes the idea to integrate multiple data sources in order to generate more reliable data compared to any individual data source. In case of sensor-generated data, data fusion combines the data of multiple sensors and the resulting more reliable and more accurate data is then processed and analyzed \cite{Kandukuri2016}, \cite{Lee2006}. On the other hand, data filtering also deals with the issue of large amounts of data by filtering uninformative data \cite{Yamato2017}. For the most part continuous monitoring will generate data which just shows that a certain machine is in a normal state \cite{Wang2016}. Therefore, data filtering models identify the useless data and only analyze the informative parts of the total amount of data, thus making the data processing more accurate and efficient \cite{Schirru2010}. However, merely the minority (14\%) of the papers directly address either data fusion, data filtering or both.

Second, in contrast to data fusion and filtering, data storage and access are attributes addressed by all papers. This is the case since every paper includes some type of monitoring which inevitably generates data that needs to be stored and made accessible for further processing. The analysis shows that for both, storage location and data access, the remote alternative is chosen by only about 20\% of the papers and mostly in combination with IoT, Industry 4.0 and cloud computing \cite{Chiu2017}, \cite{Yan2017}, \cite{Lee2006}. This fact is of interest as the remote approach seems to be more suitable for an efficient predictive maintenance system in an Industry 4.0 environment, especially with a manufacturing structure that consists of multiple production sites. Hence, the method of remote storage and access is not covered as much as expected in the recent academic literature. This issue will be discussed in Section~\ref{chap:discussion} of the present survey.

	\subsection{Evaluation}
	\label{sec:evaluation}

\begin{figure}[h]
	\centering
	\includegraphics[width=\columnwidth]{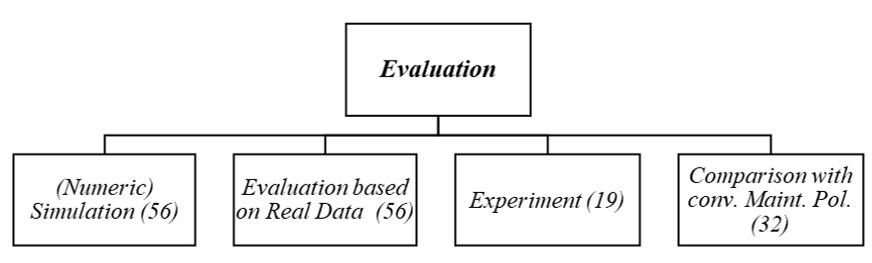}
	\caption{Overview of the Category Evaluation}
	\label{img:evaluation}
\end{figure}

Finally, the framework of the present survey for predictive maintenance is concluded by the category of evaluation. This category covers the application-oriented part of the studies. Most of the studies use some sort of data to test their predictive maintenance approach and to proof its feasibility. The following attributes were identified for this category: (Numeric) Simulation, Evaluation based on Real Data, Experiment and Comparison with Conventional Maintenance Policies. Figure \ref{img:evaluation} shows all attributes of the category and their corresponding frequency of occurrence.

Note that none of the attributes are mutually exclusive. Nevertheless, only 6 papers out of 150 choose a combination of numeric simulation and either real data or experiment. First, the numeric simulation is an evaluation based merely on simulated date, e.g. a Monte Carlo Simulation which generates data about hypothetical failures which must then be identified by the researcher's predictive maintenance approach \cite{DeSaporta2012}, \cite{Lei2015a}. More than a third (37\%) of the papers apply a simulation-based evaluation method to test their predictive maintenance techniques. Second, 37\% of the papers choose an evaluation based on real data. For this evaluation method the researchers acquire actual data from companies who monitor their machines \cite{Schmidt2017}, \cite{Elwany2008}. The data includes the monitoring data as well as information about the corresponding state of the machine, e.g. whether a failure occurred. This data can then be used to validate the accuracy and performance of the introduced predictive maintenance approach. The advantage of real data is that it allows to test the predictive maintenance approach with long term data and with data gathered from multiple large and complex machines. 

In contrast to that, the minority of papers (13\%) conducts experiments to validate their approach. Mostly, the experiments are kept very small in a sense that merely one component, e.g. bearings or a detached engine are used as experimental subjects \cite{Gebraeel2005}, \cite{Yang2002}. Thus, experiments are usually limited to small scaled test setups which makes them unsuitable to validate approaches that include multi-component systems and dependencies as well as connections with other related processes. Finally, about 20\% of the authors additionally compare their predictive maintenance approach with conventional maintenance policies as a benchmark. These comparisons are generally based on comparing total costs for different maintenance policies including predictive maintenance  \cite{VanHorenbeek2013}, \cite{Nguyen2015}. Thus, attempting to prove the superior efficiency and cost reduction opportunities of predictive maintenance.

\section{Discussion}
\label{chap:discussion}

The number of papers analyzed in the present survey is merely a fraction of the available academic literature addressing predictive maintenance. However, the analysis provides insides about which aspects the literature mostly focuses on. This discussion concentrates on four aspects that emerged during the construction and analysis of the predictive maintenance framework. First, an interesting aspect in this context is the topic of dependencies. Although 41 papers state that they take multi-component systems into account, merely 8 papers address the topic of dependencies. This implies that although predictive maintenance is more efficient when applied to an entire production system instead of single machines and when taking dependencies into consideration, the existing literature frequently focuses on a too narrow view on predictive maintenance. Many researchers either apply their approach merely to a single machine or they do not focus on dependencies of any of the three types described in subsection \ref{sec:systemsize}. Nonetheless, dependencies are severely important when attempting to design an efficient and integrated predictive maintenance system which can be applied to an entire manufacturing process. Thus, in this area seems to be a gap in the existing research and further research should focus on a broader scope for predictive maintenance systems.

Second, to date a minority of papers address modern possibilities of remote data storage and remote data access. However, with the available technology and improving internet connections for a faster transfer of data, future predictive maintenance approaches should use these technologies more severely in order to create a more centralized, networked and therefore more efficient predictive maintenance. This is especially of interest for companies with multiple manufacturing locations. When all the information can be accessed and processed at a remote location for every manufacturing site, resources can be distributed more efficiently. Furthermore, production facilities only need to install the sensor technology for monitoring and data collection purposes as the data processing is conducted remotely. Hence, connecting multiple locations to a centralized predictive maintenance system has the potential to create economies of scale. Nevertheless, since predictive maintenance is not simply scalable for any system size \cite{Nguyen2015}, the centralized system needs to take into account structural, stochastic and especially economic dependencies. Hence, with regard to an implementation of predictive maintenance in the context of Industry 4.0 remote data storage and access should come to the fore of further research.

The third aspect which might point out a gap in the existing research based on the sample of papers analyzed in the present survey is related to the category of data handling as well (see subsection \ref{sec:datahandling}). Hardly over 20 papers address the topic of data fusion and data filtering. This might be the case since a lot of research focuses on single-component systems. However, with an increasing number of sensors in use the amount of data increases and makes methods such as data fusion and data filtering necessary in order to process it efficiently. Especially in an Industry 4.0 environment with multiple integrated systems, big date becomes an issue. Predictive maintenance systems not only generate an increasing amount of data because of multiple machines or even multiple manufacturing locations but additionally due to improvements in sensor technology which will allow to monitor even more parameters then before. Wireless sensors already enable the collection of monitoring data where it was not possible to collect data before due to fragile wires. The improvements in sensor technology will support predictive maintenance systems in becoming more accurate. Nevertheless, in order to keep predictive maintenance efficient with regard to processing power usage and big data, further research is needed in the area of feasible methods of data filtering and data fusion.  

Finally, no paper in the present survey covered the topic of possibilities for small and medium sized companies. The research shows opportunities of predictive maintenance such as cost minimization, increased productivity and prolonged machine life. However, none of the papers in the present survey address the issue of how firms can implement predictive maintenance in a cost minimizing way. Thus, while the latest sensor and data processing technology might be affordable for large companies due to economies of scale, since predictive maintenance can process data of multiple sensors, small and medium sized companies are unlikely to implement predictive maintenance due to high upfront costs and complexity \cite{Carnero2006}. Nonetheless, the majority of firms are small and medium sized companies. The researchers show what is possible and do not focus on what is practical and can be implemented on a broad scale in order to boost an economies productivity and use of resources. Therefore, further research should focus on developing predictive maintenance systems that are simple and affordable to implement even for smaller companies.

To sum it up, the present survey reveals research gaps  in the context of Industry 4.0. The research often focuses on too narrow aspects instead of more broader concepts as it would be necessary in an Industry 4.0 environment. The connection of predictive maintenance to other processes such as spare part logistics should be considered in further research. Hence, predictive maintenance should be addressed as part of a larger integrated system. Furthermore, possibilities for a large scale implementation are an issue as the literature shows what is possible, but only few companies have yet successfully implemented predictive maintenance \cite{Efthymiou2012}.

\section{Conclusion}
\label{chap:conclusion}
The present survey represents a comprehensive analysis of the topic of predictive maintenance. Predictive maintenance distinguishes itself from conventional maintenance policies by the attempt to not solely detect an anomaly of a machine or component, but to predict when the failure might occur in order to efficiently schedule maintenance actions in advance. Researchers are severely interested in the concept of predictive maintenance since hundreds of papers with regard to the topic are published every year. Minimized downtime, prolonged machine life, increased productivity and reduced costs are merely a few promising prospects of predictive maintenance. The objective of the present survey is to detect and categorize a variety of aspects with regard to the comprehensive topic of predictive maintenance. For the purpose of the survey 150 papers are analyzed and categorized. The result is a data grid with 69 attributes which are clustered into the following 10 categories: Goals, Condition Monitoring, Maintenance Scope, Degradation Process, Fault Detection, System Size, Scheduling, Prognostic Techniques, Data Handling and Evaluation. These 10 categories together with their corresponding attributes built a framework for predictive maintenance.

	\subsection{Reflection of the Procedure}
	\label{sec:reflection}
For the present survey a structured literature review (SLR) is conducted. This approach is beneficial in contrast to the berry picking method as it captures all facets of the topic. Since the objective of the present survey is to categorize and structure a comprehensive scope of the research field of predictive maintenance, the SLR is the more appropriate choice. In order to construct the framework for predictive maintenance, the 150 papers are analyzed and categorized which results in 10 categories and 69 attributes each belonging to one category. Although the number of papers in the research field of predictive maintenance ranges into the thousands, the number of papers in this survey represents an informative cross subsection of the topic. Hence, the framework provides a comprehensive overview about predictive maintenance. The concept of deducing attributes, grouping them into categories and building a framework related to a specific topic is very beneficial in order to get a structured and deeper understanding of a subject and to reveal potential gaps in the existing literature. Especially when analyzing a considerable number of scientific publications.

Considering that there haven been thousands of papers published over decades with regard to predictive maintenance raised the question in which way the existing research can be classified in order to structure it and to reveal potential gaps for further research. The framework structures and classifies papers published between 1993 and 2018. The framework makes no claim of completeness, neither are the attributes within each category. Nevertheless, the framework of the present survey covers relevant aspects and facets of the topic of predictive maintenance and builds a comprehensive introduction into the research field of predictive maintenance.

The framework of the present survey shows that further research might be appropriate in certain directions of the topic. First, the effect of different types of machine and component dependencies in the context of multi-component systems. Second, remote data storage and access using IoT technology for remote predictive maintenance approaches. Third, the relevance of data fusion and data filtering due to big data issues as a result of an increasing number of monitoring sensors. Finally, the possibilities for a less complex and more affordable implementation of predictive maintenance systems even for small and medium sized companies. Note that these research gaps are presumed based on the result of the present survey which does not cover all the existing academic literature.




\bibliographystyle{elsarticle-num} 
\bibliography{predictive_maintenance_survey}





\end{document}